\documentclass[journal, 10pt, romanappendices]{IEEEtran}
\usepackage[draft, blank]{ieeefig}
\usepackage{float}
\floatstyle{plaintop}
\restylefloat{table}

\usepackage{dblfloatfix}
\usepackage{mdframed}
\usepackage{subfigure}
\usepackage{cite}
\usepackage{relsize}
\usepackage{amsopn}
\usepackage{float}
\restylefloat{table}

\usepackage{amsxtra}
\usepackage{amsmath}
\usepackage{graphicx}
\usepackage{latexsym}
\usepackage{amsfonts}
\usepackage{amssymb}
\usepackage{mathrsfs}
\usepackage{bm}
\usepackage{yhmath}
\usepackage{verbatim}
\usepackage{float}
\usepackage{xr-hyper}
\usepackage{multirow}
\usepackage{subfloat}

\usepackage{epstopdf}

\newcommand{\FR}{\mathbb{R}} 

\newcommand{\FC}{\mathbb{C}} 

\DeclareMathAlphabet{\bi}{OML}{cmm}{b}{it}
\DeclareMathAlphabet{\bcal}{OMS}{cmsy}{b}{n}
\DeclareMathAlphabet{\brmn}{OT1}{cmr}{bx}{n}

\DeclareMathSymbol{\R}{\mathalpha}{AMSb}{"52}

\newcommand{\bgamma}{\boldsymbol{\gamma}}

\newcommand{\brho}{\boldsymbol{\rho}}

\newcommand{\Fc}{\mathcal{F}}

\newcommand{\Jc}{\mathcal{J}}

\newcommand{\Lc}{\mathcal{L}}

\newcommand{\Rb}{\mathbb{R}}

\renewcommand{\o}{\omega}
\usepackage{amssymb}
\usepackage{amsfonts}

\graphicspath{{./images/}}

\DeclareMathAlphabet{\bi}{OML}{cmm}{b}{it}
\DeclareMathAlphabet{\bcal}{OMS}{cmsy}{b}{n}
\DeclareMathAlphabet{\brmn}{OT1}{cmr}{bx}{n}

\def \x{\mathbf{x}}

\def \y{\mathbf{y}}
\def \z{\mathbf{z}}

\def \a{\mathbf{a}}
\def \d{\mathbf{d}}
\def \h{\mathbf{h}}
\def \f{\mathbf{f}}
\def \W{\mathbf{W}}

\def \F{\mathbf{F}}
\def \Q{\mathbf{Q}}
\def \b{\mathbf{b}}

\usepackage{algorithm}
\usepackage{algorithmic}

\usepackage{amsthm}

\title{Deep Learning for \\ Passive Synthetic Aperture Radar}

\author{Bariscan Yonel$^1$, Eric Mason$^1$ and Birsen Yaz{\i}c{\i}$^{1,*}$, \IEEEmembership{Senior Member, IEEE}
\thanks{$^1$ Yaz{\i}c{\i}, Yonel, and Mason are with the Department of Electrical, Computer and Systems Engineering,
Rensselaer Polytechnic Institute, 110 8th Street, Troy, NY 12180 USA, E-mail: B.~Y: yazici@ecse.rpi.edu,
Phone: (518)-276 2905, Fax: (518)-276 6261.}
\thanks{* Corresponding author.}
 \thanks{This work was supported by the Air Force Office of
 Scientific Research (AFOSR) under the agreement FA9550-16-1-0234.}}

\begin{document}

\maketitle

\begin{abstract}
We introduce a deep learning (DL) framework for inverse problems in imaging, and demonstrate the advantages and applicability of this approach in passive synthetic aperture radar (SAR) image reconstruction. 
We interpret image reconstruction as a machine learning task and utilize deep networks as forward and inverse solvers for imaging. Specifically, we design a recurrent neural network (RNN) architecture as an inverse solver based on the iterations of proximal gradient descent optimization methods.
We further adapt the RNN architecture to image reconstruction problems by transforming the network into a recurrent auto-encoder, thereby allowing for unsupervised training. 
Our DL based inverse solver is particularly suitable for a class of image formation problems in which the forward model is only partially known. The ability to learn forward models and hyper parameters combined with unsupervised training approach establish our recurrent auto-encoder suitable for real world applications.


We demonstrate the performance of our method in passive SAR image reconstruction. In this regime a source of opportunity, with unknown location and transmitted waveform, is used to illuminate a scene of interest.
We investigate recurrent auto-encoder architecture based on the $\ell_1$ and $\ell_0$ constrained least-squares problem. 
We present a projected stochastic gradient descent based training scheme which incorporates constraints of the unknown model parameters.
We demonstrate through extensive numerical simulations that our DL based approach out performs conventional sparse coding methods in terms of computation and reconstructed image quality, specifically, when no information about the transmitter is available.

\end{abstract}

\section{Introduction}

Deep Learning (DL) has dramatically advanced the state-of-the-art for many problems in science and engineering. These include speech recognition, natural language
processing, visual object recognition, and many others \cite{lecun2015,Bengio2013}. In this paper, we present a novel DL framework for inverse problems in imaging and demonstrate its applicability and advantages in passive synthetic aperture radar (SAR) image reconstruction. 

In recent years, DL has drawn increasing attention in signal processing community. Several theoretical studies were conducted with the goal of connecting deep networks to established frameworks and concepts, such as wavelets \cite{mallat2016understanding, wiatowski2015mathematical}, scattering transforms \cite{mallat2016understanding, wiatowski2015mathematical, soatto2014visual2}, minimal sufficient statistics \cite{soatto2014visual}, and generative probabilistic models \cite{patel2015probabilistic}. The most notable applications of DL in signal processing were explored for compressed sensing, specifically for sparse coding and signal recovery \cite{gregor2010, rolfe2013, borgerding2016 ,mousavi2015deep, adler2016deep}.
Approaching sparse inverse problems with deep neural networks that mimic the coordinate descent (COD) and iterative shrinkage thresholding algorithm (ISTA) were first proposed in \cite{gregor2010}. This approach was then extended to the approximate message passing (AMP) algorithm, and was shown to have improved performance over the ISTA based counter part \cite{borgerding2016}. 
For both methods, the neural network is formed by unfolding the iterative optimization method, yielding a \textit{recurrent neural network} (RNN) model. 
The connection between optimization and DL has been further investigated in recent studies for learning problem specific gradient descent parameters \cite{andrychowicz2016learning}, and estimating priors for inference from data \cite{putzky2017recurrent}.
For image reconstruction, DL with convolutional neural networks has been recently described for problems in which the underlying normal operator is Toeplitz. This method was shown to outperform total variation regularized reconstruction for sparse view X-Ray computed tomography \cite{jin2017deep}.

With the exception of \cite{jin2017deep}, existing studies explore applications of DL in generic signal recovery problems. In this study, we are primarily interested in exploring DL framework for inverse problems in imaging which can be categorized into three classes: forward modeling, that is modeling the relationship between the quantity of interest and measurements; inversion, that is forming an image of the quantity of interest from measurements; and finally design of algorithms for computationally efficient forward and inverse solvers. We postulate that DL framework can be exploited to address all three aspects of inverse problems in imaging.

We start with a review of key concepts and tools in DL and present the conceptual evolution from conventional to DL based machine learning algorithms. While conventional machine learning algorithms can be viewed as a two-layer process, DL based approach involves multiple hidden layers inserted in between the two layers and a non-linearity in each layer. We next provide an interpretation of forward modeling and image reconstruction as machine learning tasks and present extensions from the conventional two-layer linear processing to multi-layer non-linear processing as hidden layers. As a result, deep network becomes a non-linear forward model potentially capturing the relationship between physical measurements and quantity of interest more accurately than its linear counterparts. 
In addressing the inversion problem, we extend the conventional two-layer processing involving filtering and backprojection to DL based approach by inserting multiple hidden layers in between filtering and backprojection steps. We next motivate the choice of network non-linearity by non-Gaussian prior models in a
Bayesian formulation of the image reconstruction problem and the solution of the resulting minimization problem via numerical optimization algorithms. Specifically, we design an RNN architecture as an inverse solver based on the iterations of proximal gradient descent optimization methods.
In this approach, network bias becomes backprojected data, network weights serve as an image domain filter and non-linear activation function is chosen based on prior information. 
We further adapt the RNN architecture to image reconstruction problems by adding an additional layer mapping and transforming the network into a recurrent auto-encoder \cite{rolfe2013}. Unlike the method used in \cite{jin2017deep}, our network allows for unsupervised learning using data directly. As a result, the performance of our DL based inverse solver is not upper bounded by the quality of training images generated by conventional methods.

Our DL based inverse solver is particularly suitable for a class of image formation problems in which the forward model is only partially known. Such problems can arise due to simplifying assumptions, unknown parameters, and uncertainties in the underlying physical/sensing process. The ability to learn or refine forward models, hyper parameters combined with unsupervised training approach establish our recurrent auto-encoder suitable for real world applications. 
The learned network reduces computation by requiring a smaller number of layers than the number of iterations in standard sparse coding. 
Our approach is distinct from \cite{borgerding2016,gregor2010,rolfe2013} in which DL is utilized to learn parameters and network weights based on random model initialization. Instead, we use physics based modelling as initialization for an auto-encoder structure where the forward model is restricted to be the same in the encoder and the decoder, perform unsupervised training with raw data, and obtain image estimate after the encoder.

To demonstrate the applicability and performance of DL based imaging, we consider passive synthetic aperture radar (SAR) image formation problem.
Passive SAR uses existing sources of opportunity to illuminate a scene of interest and deploys passive receivers to measure the backscattered signals \cite{Hack12, Davidowicz12,Palmer13,Yarman08,LWang12}.
Using illuminators of opportunity results in many challenges such as limited bandwidth, non-ideal transmitted waveforms, and uncertainty in knowledge of the transmitter location and transmitted waveforms. Existing approaches to the problem either make strong assumptions to reduce dependence on the transmitter location \cite{Yarman08,Wacks14}, assume direct-line-of-sight to the transmitter \cite{Hack12, Davidowicz12,Kulpa12}, or are computationally inefficient \cite{Mason2015}.
We demonstrate the power of DL by developing a novel approach to passive SAR, overcoming these shortcomings.
Specifically, we do not assume any knowledge of the transmitter location or properties in image reconstruction and learn the forward model that depends on these parameters.
We then use the DL network to learn the missing transmitter information which allows for efficient reconstruction of arbitrary scenes. We demonstrate that application of our method to passive SAR imaging results in better background suppression, target localization, reduced reconstruction error and higher image contrast than that of ISTA or iterative hard thresholding algorithm (IHTA), requiring less layers (or iterations) resulting in significantly reduced computation. 

The rest of the paper is organized as follows: in Section \ref{Sec:DL_backround}, we provide a brief overview of deep learning (DL). Next, in Section \ref{Sec:DL_inverse}, we establish the connection between machine learning and inverse problems in imaging and present optimization based imaging. In Section \ref{Sec:SARDL} we introduce the passive SAR imaging problem, address it within DL framework, and discuss our network design choices and training details. In Section \ref{Sec:NumericalSim}, we demonstrate the performance of our method with numerical results. Section \ref{Sec:Concl} concludes the paper.


\section{Deep Learning}\label{Sec:DL_backround}

Deep Learning falls into a class of machine learning methods known as representation learning. These methods excel in extracting features from the data automatically, bypassing the hand-crafting process of feature design. These algorithms are characterized by a cascade of many ``layers" of element-wise non-linear operations. In the most general sense, such algorithms are categorized as artificial neural networks (ANNs). In order to extend Deep Learning applications to image formation, we first give an overview of fundamental concepts and building blocks of DL-based machine learning and ANNs.

We begin by introducing key concepts for DL, proceed with a comparison of DL to conventional machine learning methods, present some fundamental architectures of the framework. In the next section, we discuss how to formulate inverse problems in imaging in DL framework. 

\subsection{Fundamental Concepts}

The cornerstone of the field of machine learning is the \textit{perceptron} classifier.
This is a fundamental computation unit, parameterized by a weight vector $\mathbf{a} \in \mathbb{C}^N$ and a bias term $b \in \mathbb{C}$.
The operation of the perceptron is merely the projection of the input vector $\mathbf{f} \in \mathbb{C}^N$ on $\mathbf{a}$ via dot product, summation with the bias $b$, followed by the unit step function $\sigma$. 
The perceptron, as a classifier, decomposes the input space into two decision regions lying above and below the hyperplane with the normal vector $\a$, displaced from origin by $b$ units. The class assignment is performed by the unit step function $\sigma$ on the output $\a \cdot \f + b$.

An arbitrary choice of the non-linear function $\sigma$ generalizes the concept of perceptron to an artificial \emph{neuron}. 
The function $\sigma$ is referred to as the \emph{activation function}. Parallel implementation of multiple neurons constitute a processing block, namely a \emph{layer} \cite{Goodfellow2016}. A layer is an an affine mapping of the input $\f \in \Omega_{\f} \subseteq \mathbb{C}^{N}$, 
 followed by an elementwise non-linear warping:
\begin{equation}\label{eq:layerOutput}
  \tilde{\f} =  \sigma ( \mathbf{Af + b} )
\end{equation}
where rows of the matrix $\mathbf{A}\in \mathbb{C}^{M \times N}$ are weight vectors of the underlying neurons, and $\mathbf{b}\in \mathbb{C}^{M}$ is vector of corresponding biases. The output $\tilde{\mathbf{f}} \in \mathbb{C}^M$ is called a \emph{representation} of the input $\mathbf{f}$. A representations lies in a ``feature space", the space resulting from the transformation of the input space $\Omega_{\f}$ by the layer. Serial connection of multiple layers constitutes a ``model" or a network ``architecture" to perform classification or regression on the input data.

\subsection{Conventional Machine Learning vs Deep Learning}

The general task of machine learning is to estimate model parameters to produce descriptive representations and to identify decision regions at the output space.
The parameter estimation procedure is referred to as learning, or training of the model.
In principle, machine learning methods are inductive, i.e., the learning takes place by experience or examples. The learning procedure can be either supervised or unsupervised depending on whether the training set includes ground truths or not.



\begin{figure}[!ht]
\label{fig:layer}
\centering
\subfigure[]{
\includegraphics[scale=0.23]{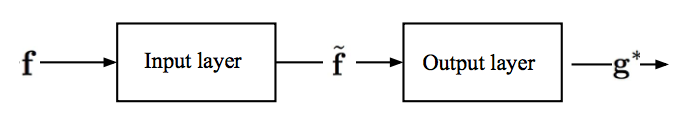}
\label{fig:2layer_arch}
}
\hspace*{-0.72cm}
\subfigure[]{
\includegraphics[scale=0.30]{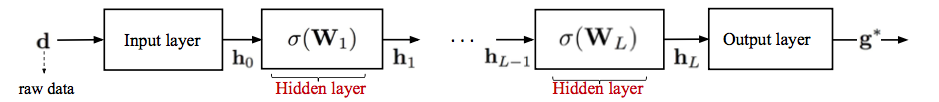}
\label{fig:deep_arch}
}
\caption{(a) Conventional machine learning vs. (b) Deep Neural Network with additional layers between input and output layers. In conventional machine learning the input $\f$ is a hand-crafted feature whereas in Deep Learning it is raw data $\d$.}
\end{figure}

In conventional machine learning, most methods are 2-layered architectures as shown in Fig. \ref{fig:2layer_arch}. The input to the first layer is a hand-crafted feature. The result of the first layer is a new representation in a feature space. The ``output layer'' maps this new representation to decision regions. It is important to note that in conventional machine learning, features are hand-crafted from raw data specific to a problem in hand. Therefore, feature engineering is a critical part of classifier design.


Unlike conventional machine learning, DL does not require feature engineering. In DL, useful features can be learned directly from raw data, $\d \in \Omega \subseteq \FC^N$,
in a hierarchical manner. This is achieved by expanding the conventional shallow model into a chain of many processing units by inserting layers in between the standard 2-layer structure as shown in Fig. \ref{fig:deep_arch}. These new layers are referred to as ``hidden'' layers.
\subsection{Forward Propagation in DL}

With the additional layers, the model produces a new representation at each layer of the output generated at the previous layer, resulting in a hierarchical representation of the input. As a result, the input is represented in increasingly abstract feature spaces. 
The output at the end of the $k^{th}$ layer can be written as:
\begin{equation}\label{eq:singlelayeroutput}
\h_{k}  =  \sigma ( \mathbf{A}_{k}  \h_{k-1} + \b_{k} ).
\end{equation}
Let $\phi : \Omega_L \rightarrow \Gamma $ be a mapping from the feature space $\Omega_L$ defined by the range of the hidden layer to the output space $\Gamma$. 
Redefining $\d = [\d, \ 1]^T$ and $\W_k = [\mathbf{A}_k, \ \b_k]$, $k=0,...,L$, then, network output $\mathbf{g}^* \in \Gamma$ is given by:
\begin{equation} \label{eq:layeroutput}
\mathbf{g}^*  =  \phi\left(\W_L \sigma(\W_{L-1} ...  \sigma(\W_{1} \sigma(\W_{0} \mathbf{d})\big)\right).
\end{equation}
The process of mapping the raw data $\d \in \Omega$ to the output space, $\Gamma $, is referred to as the \emph{forward propagation}.
As a direct consequence of composition of transformations performed by successive layers, deep networks gain more expressive power and the ability to represent more complex mappings. These sophisticated transformations are generally explained in terms of the universal approximation theorem or probabilistic inference, and can theoretically approximate any function \cite{Hornik1991}. The weights provide a linear parametrization of the network operation and the activation function introduces the capacity to approximate complex, non-linear mappings between input and output spaces.


The mapping of the output layer $\phi (\cdot)$ is chosen in a problem-specific manner. This establishes DL as a goal driven method. Network parameters are learned to construct feature spaces that are relevant to a specific problem in hand.

\subsection{Learning in DL}

Let $\mathcal{L}(\theta) : \Omega \rightarrow \mathit{Y}$ be the mapping between the input and output spaces where
\begin{equation}\label{eq:NetworkOperator}
  \mathcal{L}(\theta) [\mathbf{d}] = \mathbf{g^*}
\quad \quad \theta = \{\mathbf{W}_k\}_{k = 1}^{L}.
\end{equation}
$\mathcal{L}(\theta)$ is referred to as the network operator, with network parameters $\theta$. In DL, learning involves estimating $\theta$ with respect to a figure of merit given a set of training data $\left\{ \mathbf{d}_1, \mathbf{d}_2, \cdots,  \mathbf{d}_T \right\}$ and corresponding ground truth data set $G = \left\{ \mathbf{g}_1, \mathbf{g}_2 \cdots,  \mathbf{g}_T \right\}$. A commonly used figure of merit is the $\ell_2$ error,
\begin{equation}\label{eq:L2Error}
  \Jc_{G}[\theta] =  \frac{1}{2T} \sum\limits_{n=1}^T \|  \Lc(\theta)[\mathbf{d}_n] - \mathbf{g}_n \|^2_2.
\end{equation}
The minimization of $\Jc_{G}[\theta]$ is typically a high-dimensional and non-convex optimization problem, often with many saddle points and local minima \cite{Goodfellow2016, lecun2015}.
Most widely used approach in addressing this optimization problem is the gradient descent method producing the following updates:
\begin{equation}\label{eq:updates}
\theta^{l+1} = \theta^{l} - \eta_l \nabla_\theta\Jc_{G}[\theta^l],
\end{equation}
where $l$ denotes the $l^{th}$ update and $\eta_l$ is called the learning rate. 
The method of updating the network parameters with the gradient descent principle is called \emph{backpropagation}, which is the fundamental way of learning in deep networks.

The method by which the gradient term, $\nabla_\theta\Jc_{G}[\theta^k]$, is computed over the training data leads to different parameter update schemes. For large training sets, the most common update method is the stochastic gradient descent(SGD). In SGD, at each iteration, the update term is estimated by averaging the computed values of the gradient over a small subset of the training set. Going over the entire training set in computing updates completes an ``epoch''. 
Alternative methods to SGD are most notably the batch and on-line updates. The batch update averages the computed gradients over the entire training set, hence performs one update per epoch, and on-line update computes the gradient for each datum and updates weights of the model sequentially.
More sophisticated optimizers that are variants of SGD such as, AdaGrad \cite{duchi2011adaptive}, Adam \cite{kingma2014adam}, RMSProp \cite{tielemanrmsprop}, and forward-backward splitting optimizers \cite{duchi2009efficient}, as well as second order optimization methods \cite{martens2010deep} have been proposed in the literature for backpropagation to improve training deep networks.

%


\subsection{Generic Architectures in ANNs}
Several different network architectures have been developed through the development of ANN in DL literature.
Some of the core architectures can be listed as deep feedforward neural networks (FFNNs), convolutional neural networks (CNNs), recurrent neural networks (RNN) and auto-encoders \cite{lecun2015, Bengio2013, Goodfellow2016}.


In FFNNs, each layer is an independent processor, hence parameterized uniquely. Therefore, at each layer a different transformation is learned. 
CNNs employ Toeplitz matrices, hence convolutions, as network weights. They are suitable to model sparse, localized interactions and are widely used in computer vision tasks. RNNs are highly structured forms of deep FFNNs, in which all layers perform an identical transformation. These identical transformations can be interpreted as a recurring state update, in which each layer represents a time step. 

These basic architectures are used as building blocks of more complex deep networks. 
One such class of architectures is the auto-encoders. Auto-encoders consist of two components, an encoder, and a decoder. Each component could be any ANN architecture. The decoder block synthesizes the input from the output of the encoder, and a desired representation is extracted from the intermediate stage between the two components. An encoder-decoder pair is learned from data in an unsupervised manner.

\section{Deep Learning for Inverse Problems in Imaging}\label{Sec:DL_inverse}

Deep Learning framework has been traditionally developed for machine learning tasks. In this section, we introduce an interpretation of inverse problems in imaging as machine learning tasks and explore perspectives, concepts and tools offered by DL in addressing imaging problems.

Inverse problems in imaging can be categorized into three components: forward modeling, that is modeling the relationship between the quantity of interest and measurements;  inversion, that is forming an image of the quantity of interest from measurements; and finally design of algorithms for computationally efficient forward and inverse solvers. DL framework may offer advantages and play an integral unifying role in addressing all three components of imaging. Specifically, we postulate that Deep Learning theory can be exploited for the following key tasks: \emph{i}) learn or refine forward models; \emph{ii}) develop image reconstruction methods for ill-posed inverse problems; and \emph{iii}) design computationally efficient algorithms.


\subsection{Deep Network as a Forward Solver}
In many inverse problems of imaging, forward modeling is an art of compromise between accuracy and tractability. As a result forward models often rely on simplifying assumptions, approximations and include uncertainties. For example, in many electromagnetic wave based imaging problems, it is not computationally tractable to use Maxwell's equation. Instead, one would use wave equation and further simplify the underlying non-linearity under Born or single scattering assumptions to arrive at a linear model between quantity of interest and physical measurements. In some problems, the resulting linear model is further simplified under far-field approximations to obtain the Radon transform. The resulting mathematically idealized model may rely on simplified system, imaging and medium parameters which may be at best approximate in practice.

DL offers the possibility of estimating a non-linear forward model, when forward model is implemented in forward propagation. We can then interpret physical measurements as representations of image/physical quantity of interest to be learned in measurement space. Let
\begin{equation}\label{eq:ForwardModel}
      d = \mathcal{F}[\rho]+noise
    \end{equation}
where $\rho : \FR \times \FR \rightarrow \FR^+ \cup \{0\}$ is the quantity of interest, $d : \FR^l \rightarrow \FC$ denotes physical measurements, and $\mathcal{F}$ is the forward model.

In conventional forward modeling $\mathcal{F}$ is often simplified to a linear model and the observed measurements can be interpreted as an output of a 2-layer network as in Fig. \ref{fig:2layer_arch}, in which the input layer performs the linear operation governed by $\mathcal{F}$ and the output layer is merely an identity operator or an additive noise. 

DL framework, on the other hand, suggests to increase the number of layers and uses affine transformations followed by a non-linearity in each layer as depicted in Fig. \ref{fig:deep_arch}. Inserting $L$-hidden layers results in the following non-linear model between the quantity of interest $\rho$ and measurements $d$:
\begin{equation}\label{eq:DL-ForwardModel}
      d = \tilde{\sigma}(\tilde{\mathcal{W}}_L \tilde{\sigma}(\tilde{\mathcal{W}}_{L-1} ...  \tilde{\sigma}(\tilde{\mathcal{W}}_{1} \tilde{\sigma}(\tilde{\mathcal{W}_0} [{\rho}])\big)\big).
    \end{equation}
    
Given a set of training data $\{\rho_{t1}, ..., \rho_{tT}\}$ and corresponding measurements $\{d_{t1}, ..., d_{tT}\}$, we can then use DL's backpropagation step to learn network parameters, $\{\tilde{\mathcal{W}}_{k}\}_{k=0}^{k=L}$. Since this is a highly non-convex problem, we can choose to model input layer by the linear model $\mathcal{F}$ to obtain a relatively good initialization to the backpropagation algorithm. 


\subsection{Deep Network as an Inverse Solver}\label{Ssec:DLInvSolv}
%
Inversion or image formation can be implemented as forward propagation in DL framework. We can then interpret inversion as a process of finding a desired representation of the physical measurements in image space. This leads to a formulation of image reconstruction as representation learning in DL framework.


Given a linear forward model as in (\ref{eq:ForwardModel}), we can implement conventional image reconstruction in two-steps involving backprojection and filtering:
\begin{equation}\label{eq:LinearInvModel}
      {\rho^*} = \mathcal{K}[d] := \mathcal{F}^\dag \mathcal{Q}[d]
    \end{equation}
where $\mathcal{F}^\dag$ denotes the backprojection or adjoint operator and $\mathcal{Q}$ is a filter that can be designed with respect to a variety of criteria.

The backprojection-filtering operation in (\ref{eq:LinearInvModel}) has an analogous interpretation to 2-layer conventional machine learning in which the input layer comprises of backprojection and the output layer comprises of filtering. Backprojection operator produces a representation of measurements in the range of the adjoint operator and filtering refines this representation with respect to a criterion such as suppressing singularities due to noise or sharpening edges. \footnote{The filtered-backprojection operator can be also interpreted as a 2-layer network with the order of filtering and backprojection switched.} 

DL suggests to introduce multiple hidden layers between the input and output layers in which each layer produces a new representation progressively approaching to a desired image. Mathematically, we can then express the reconstructed image as follows:
\begin{equation}\label{eq:DL-InverseModel}
{\rho}^* = \sigma(\mathcal{W}_{L+1} \sigma(\mathcal{W}_{L} ...  \sigma(\mathcal{W}_{1} \sigma(\mathcal{W}_0 [{d}])\big)\big).
\end{equation}
The critical advantage offered by Deep Learning framework is the learning step, hence backpropagation. Given a set of training data $\{\rho_{t1}, ..., \rho_{tT}\}$ and corresponding measurements $\{d_{t1}, ..., d_{tT}\}$, we can estimate network parameters, $\{\mathcal{W}_k\}_{k=0}^{k=L+1}$. Such an approach unifies the forward modeling and inversion and can be regarded as a model-free inversion method. However, since backpropagation algorithm is a high dimensional, non-convex optimization problem, we can set input layer weights as $\mathcal{W}_0 = \mathcal{F}^\dag$ in Fig. \ref{fig:deep_arch} to initialize optimization at a relatively good estimate. The output layer can be a linear/non-linear filter chosen with respect to a criterion such as noise suppression or detection of edges.


\subsection{Bayesian and Optimization Inspired Deep Learning based Imaging}\label{Ssec:DL_opt}

A key question that arises is: how the network activation function, $\sigma$, can be designed or chosen in using DL as an inverse solver? We propose that $\sigma$ can be chosen or designed based on \emph{a priori} information of the unknown image. Depending on the specific problem in hand, $\sigma$ may be known and can be specified entirely or a functional form of $\sigma$ may be known and its unknown parameters can be learned in backpropagation along with the rest of the network parameters.

The role of activation function for promoting a priori information can be understood by considering a Bayesian formulation of the imaging problem and the solution of the resulting optimization problem via numerical optimization algorithms.

Choosing appropriate basis functions for the image and data domains, we now replace $\rho$, $d$ and $\mathcal{F}$ with their finite dimensional versions of $\brho \in \mathbb{R}^M$, $\d \in \mathbb{C}^N$ and $\F \in \mathbb{C}^{N \times M},$ and consider the following constrained least-squares problem\footnote{Although many of the subsequent results can be extended to infinite-dimensional spaces, we restrict ourselves to finite dimensional setting to avoid technicalities for the rest of the paper.}:
\begin{equation}
\label{eq:OptInvModelDisc}
\brho^* = \underset{\brho}{\text{Argmin}} \ \frac{1}{2}\|\F\brho - \d \|_2^2 + \lambda \Phi(\brho)
\end{equation}
where $\Phi$ is a regularization term promoting \emph{a priori} information on $\brho$ and $\lambda$ is a scalar trading-off between the data-fidelity and prior information. While Gaussian prior models promote analytic linear reconstruction methods, Bayesian problem formulation along with numerical optimization algorithms allow incorporation of more general, non-Gaussian priors and development of non-linear reconstruction algorithms. 
One such class of optimization algorithms is the \emph{forward-backward splitting} \cite{combettes2009proximal}. These algorithms consist of a gradient step over the smooth $\ell_2$ data-fidelity term, called the forward step, followed by a backward step by the scaled \emph{proximity operator} of the convex penalty $\Phi$: 
\begin{equation}\label{eq:NetworkLayer_PGD}
\brho^{k+1} = \mathcal{P}_{\alpha \lambda \Phi} ((\mathbf{I}-\alpha \mathbf{F}^H \mathbf{F}) \brho^k + \alpha \mathbf{F}^H \d)
\end{equation}
where $\brho^k$ is the $k^{th}$ iterate, $\alpha$ is a properly chosen step size of the gradient descent to ensure convergence and $\mathcal{P}_{\alpha \lambda \Phi}$ is the proximity operator of $\alpha \lambda \Phi$ defined as \cite{combettes2009proximal}:
\begin{equation}\label{eq:ProxOp}
\mathcal{P}_{\alpha \lambda \Phi}(\brho) = \underset{\y}{\text{Argmin}} \ \frac{1}{2}  \| \brho - \y \|_2^2 + \alpha \lambda \Phi(\y).
\end{equation}
When $\Phi$ is the indicator function for the convex set $C$, the proximity operator reduces to the projection operator. As such, proximal gradient decent algorithms are generalization of projection gradient descent algorithms. The proximity operators exist for a handful of non-convex penalties, such as the $\mathit{l}_0$ norm penalty \cite{blumensath2008iterative} under the condition that the minimizer of (\ref{eq:ProxOp}) is unique. An extensive survey and table of proximity operators for convex penalties can be found in \cite{combettes2009proximal}.

\begin{figure*}\label{fig:RNNdiag}
\centering
\includegraphics[scale=0.8]{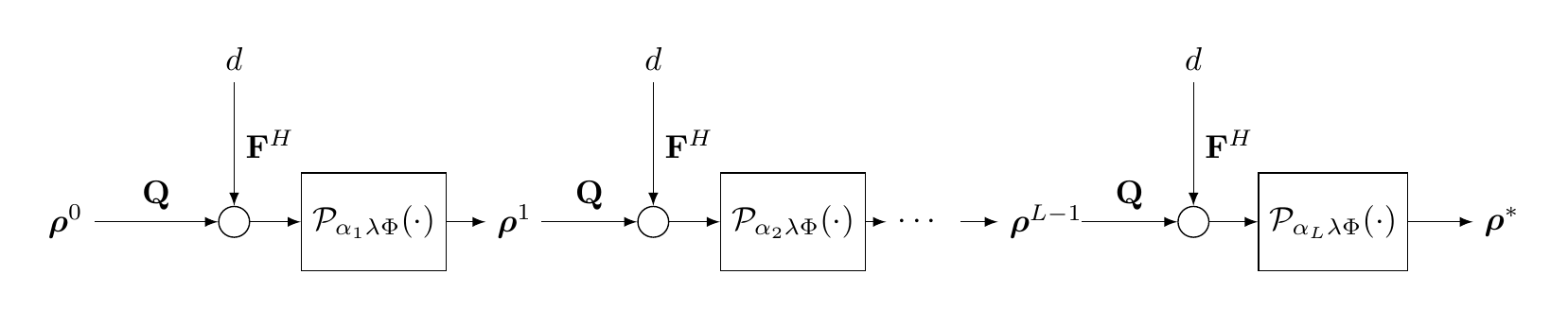}
\caption{The recurrent network architecture. The linear stages $\mathbf{Q}$ and $\b = \mathbf{F}^H \d$ are represented as arrows. The non-linear activation functions are represented as boxes. At each layer, the linear gradient descent step is followed by the element-wise proximity operator with parameter $\alpha \lambda$ shown in equation (\ref{eq:NetworkLayer_PGD}). Note that $\brho^0 = \mathbf{0}$ conventionally.}
\end{figure*}

Following the principles we lay out in Section \ref{Ssec:DLInvSolv}, we view the deep structure and generation of sequence of representations analogous to an iterative algorithm to solve an optimization problem for a fixed number of iterations \cite{gregor2010, borgerding2016}.
We unfold the iterations of (\ref{eq:NetworkLayer_PGD}) and design a DL inverse solver based on the optimization algorithm in (\ref{eq:NetworkLayer_PGD}) in which $\brho^{k}$ iterates are now the representations produced at each layer,
\begin{equation}\label{eq:WeightM}
  \Q = \mathbf{I}-\alpha \mathbf{F}^H \mathbf{F}, \quad
  \b = \alpha \mathbf{F}^H \d,
\end{equation} 
where $\Q$ is the weight matrix, and $\b$ is the bias vector of the layers, and $\mathcal{P}_{\alpha \lambda \Phi}$ serves as the activation function of the network.
The bias vector $\b$ becomes the backprojected image and the weight matrix $\Q$ is an image domain filter. Since the parameters of the network are identical at each layer, the solver for the optimization problem naturally takes the form of a \emph{recurrent neural network}. Fig. \ref{fig:RNNdiag} illustrates the architecture of the resulting RNN.

Clearly, the Bayesian problem formulation and proximal gradient descent algorithms inspire DL based solvers and provide insightful interpretations of network parameters. One of the conditions for such an interpretation to hold is that the penalty function $\Phi$ must have a closed form, element-wise proximity operator. Nevertheless, we can modify/approximate proximity operators to satisfy such a condition or to achieve certain objectives. We can further relax network structure and learn different image domain filters, backprojection operators and different step sizes, $\alpha$, at each layer as shown in Fig. \ref{fig:RNNdiag}.

The coupling of DL and optimization frameworks for image reconstruction offers advantages to both approaches.
DL offers a new framework for simultaneous image reconstruction and estimation of the underlying unknown parameters. In DL framework, the minimization alternates between different variables, namely the unknown image and network parameters, by alternating between forward propagation and backpropagation. In optimization framework, on the other hand,  the minimization typically alternates between different variables at every iteration.
In DL framework, backpropagation improves the performance of a task specific forward propagation, promoting DL as a task driven optimization. Given the interpretation of the underlying network parameters provided by the Bayesian approach and optimization algorithms, DL can address imaging problems in which complete knowledge of the forward model or \emph{a priori} information is not available. The network activation functions can be viewed as non-Gaussian priors for inference with deep models, whereby representations get warped to new feature spaces between layers. Bayesian point of view provides valuable perspective on how to determine the choice of activation function for deep networks for image reconstruction tasks.

\section{Passive Synthetic Aperture Imaging}\label{Sec:SARDL}

In this section, we introduce the passive synthetic aperture imaging problem and in the subsequent section demonstrate the applications of the methods introduced in Section \ref{Sec:DL_inverse} for passive synthetic aperture radar imaging.

A passive SAR system uses transmitters of opportunity as an illuminator and a moving receiver to image a scene of interest. %
Passive radar has been the subject of intense research in recent years due to proliferation of transmitters of opportunity and other advantages offered by passive systems, such as efficient use of electromagnetic spectrum, increased stealth capability, reduced cost and deployment flexibility among others. For an account of recent work on passive radar, see \cite{Malanowski09, Hack12, Davidowicz12, Kulpa12, Stinco13, Palmer13, Malanowski14, Wang10, wang12, LWang12, Wacks14, Mason2015}. 

At a system level, passive imaging methods can be broadly categorized into two classes: passive coherent location (PCL) and interferometric passive imaging. PCL requires two antennas at each receiver location, one directed to a transmitter of opportunity and another one directed to the scene of interest. The signal received directly from the transmitter of opportunity is used to backproject the signal scattered from the scene via matched filtering. This method requires two antennas at each receiver location, direct line-of-sight to the transmitter of opportunity and high signal to noise ratio (SNR) for the signal received directly from the transmitter. An alternative to PCL is the interferometric passive imaging. This method uses two or more sufficiently far apart receivers deployed on the same or different platforms. The signals at different receiver locations are correlated and backprojected based on time or frequency difference of arrival to form an image of the scene \cite{Yarman08, wang12, Wacks14}. 
The latter technique does not require direct-line-of-sight to a transmitter, high SNR, or the knowledge of transmitter location. However, it is limited to imaging widely separated point scatterers. More recently, an alternative method based on low-rank matrix recovery using interferometric measurements has been developed \cite{Mason2015}. 
However, computational requirements of this method precludes its applicability in reconstructing realistic sized images.

In the rest of this section, we formulate a passive SAR imaging method based on DL that is neither PCL nor interferometric in nature. We assume that a transmitter of opportunity illuminates a scene of interest. Neither the location of the transmitter nor the transmitted waveform are known a priori. A single antenna receives scattered signal from the scene as it moves along an aperture. In such a set-up, an image of the scene cannot be formed by neither classic bistatic imaging techniques \cite{Yarman08} nor PCL or interferometric techniques as the forward model is not fully known. Our objective is to address the image reconstruction in the DL framework introduced in Section \ref{Sec:DL_inverse}.

We now describe the SAR received signal model and DL based SAR image reconstruction method. In Section \ref{Sec:NumericalSim}, we present numerical simulations to demonstrate the performance of DL based SAR imaging.

We reserve the variables $\x$ and $\z$ to denote the location of scatterers in $\mathbb{R}^3$. Let $\bi  x = [x_1,x_2] \in \Rb^2$ denote the two-dimensional position of a scatterer on the ground plane and $\psi:\Rb^2 \rightarrow \Rb$ be the ground topography. Then, a scatterer is located at $\x = [\bi x,\psi(\bi x)] \in \Rb^3$.
Let $\rho(\bi x)$ denote the reflectivity of the scene.

Using scalar wave equation and under the Born approximation, we can model the received signal $d$ as follows \cite{Yarman08}:
  \begin{equation}
d(\o,s) \approx \Fc[\rho](\o,s) := \int \mathrm{e}^{-\mathrm{i}\frac{\o}{c_0}R(s,\bi x)} A(\o,s,\bi x) \rho(\bi x) d\bi x
\label{eq:FIO_forward_model}
\end{equation}
where $s \in [s_1,s_2]$ is the slow-time variable parametrizing the location of a moving antenna, $\o\in[\o_1,\o_2]$ denotes the fast-time temporal frequency, $c_0$ denotes the speed of light in free-space and $A(\o,s,\bi x)$ is a slow-varying function of $\o$ that depends on transmitted waveforms, antenna beam patterns and geometric spreading factors.
$R(s,\bi x)$ is the bistatic range given by
\begin{equation}
R(s,\bi x) = |\bgamma_T - \x| + |\bgamma_R(s)-\x|
\label{eq:mono_range}
\end{equation}
where $\bgamma_R(s)$ and $\bgamma_T$ denote the receiver and transmitter locations, respectively. Note that the transmitter is stationary, but its location $\bgamma_T$ is not known, neither is the transmitted waveforms nor the transmitted antenna beam patterns. As a result the forward model $\mathcal{F}$ is only partially known.
If the forward model $\F$ is known, a bistatic SAR image with good geometric ﬁdelity can be formed by a two-layer filtered-backprojection type operation \cite{Yarman08}.



\section{DL based Passive SAR Image Formation}\label{Sec:DLPassive}

In this section, we formulate passive SAR image formation in DL framework and describe the network architecture and training, forward propagation and backpropagation. In each of these topics, we discuss network properties specific to the passive radar image reconstruction. 

\subsection{Network Architecture and Training}
We now consider the finite dimensional versions of $\rho \rightarrow \brho \in \mathbb{R}^M$, $d \rightarrow \d \in \mathbb{C}^N$ and $\Fc \rightarrow \F \in \mathbb{C}^{N \times M}$. 
We implement two $L-$layered RNNs mimicking two proximal gradient methods in forward propagation, namely \textit{iterative hard thresholding algorithm} (IHTA) and \textit{iterative shrinkage thresholding algorithm} (ISTA) and choose the network activation functions based on the proximity operators of the $\ell_0$ and $\ell_1$ norm penalties. In such a network, the weight matrix and bias are fixed across all layers. The bias has the interpretation of the backprojected received signal and the weight has the interpretation of an image domain filter when the reconstruction is formulated as an optimization problem and addressed by proximal gradient descent type algorithms. The proximity operator of the optimization serves as the network's activation function. The number of layers, $L$, in the network simply corresponds to the number of iterations in the numerical optimization.

\begin{figure}\label{fig:autoencoder}
\centering
\includegraphics[scale=0.5]{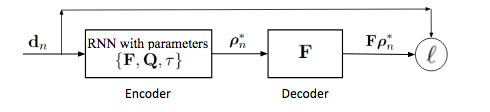}
\caption{The Auto-Encoder architecture for SAR image reconstruction. The encoder is an RNN with network parameters $\F, \Q$ and $\tau$ producing a representation, i.e., an image $\brho^*_n$. The decoder is the forward model $\F$. $\ell$ denotes a mismatch function between the synthesized measurement $\F\brho^*_n$ and actual measurement $\d_n$. This architecture allows an unsupervised learning mechanism for the network parameters.}
\end{figure}

A common approach in DL for training deep networks is by supervised learning using a training set with ground truth information. However, in the context of image reconstruction, SAR images with ground truth information may not be available. Even if such as set is available, the performance of the DL based inverse solver would be upper bounded by the quality of reconstructed SAR images in the training set. Therefore, we choose an \emph{unsupervised} learning for DL based image reconstruction and estimate network parameters from a set of training data containing solely physical measurements, $\{\d_1,...,\d_T\}$. Such an unsupervised learning is achieved by inserting an additional linear stage after the RNN to synthesize measurements from the image estimate and comparing the actual measurements with the synthesized measurements by means of a mismatch function. This linear stage is naturally the SAR forward projection model. 
Thus, our DL inverse solver becomes an \textit{auto-encoder} with SAR forward model serving as the decoder and the RNN serving as the encoder. Fig. \ref{fig:autoencoder} depicts the architecture of our auto-encoder network. The auto-encoder refines the unknown image domain filter $\Q$ and forward model $\F$ so that the measurements synthesized from the reconstructed image match the true physical measurements, driving the RNN to generate more accurate images. 

In addition to $\F$ and $\Q$, we also learn $\tau$, the parameter associated with the network activation function, by minimizing the following functional using a training data set $D = \{\d_1,...,\d_T\}$:
\begin{equation}\label{eq:Mismatch}
           \mathcal{J}_{D}[\F, \Q, \tau] = \frac{1}{T}\sum_{n=1}^T \ell(\F\brho^*_n(\F,\Q, \tau), \d_n)
         \end{equation} 
where $\brho^*_n$ is the image estimate produced by the forward propagation using the training data $\d_n$ and $\ell$ is an appropriately chosen mismatch function. Note that the loss function in (\ref{eq:Mismatch}) can be further enhanced by including appropriate constraints on the network parameters.
Such an unsupervised learning approach establishes our network suitable for real world applications, as the learning is performed using a mismatch function defined on the measurement space without the need for ground truth SAR images.

While we have discussed the general network and training scheme for passive SAR imaging, there are several design choices and specific network properties that impact the network performance. In the following subsections, we present the details of the RNN, network derivatives for SGD based backpropagation and justifications for the design choices. 

\subsection{Forward Propagation for Passive SAR}\label{Ssec:ForProp_PSAR}


A key component of the RNN network that needs to be determined is the domain of representations, $\brho^k$, and the encoder output, ${\brho}^*$.
The entries of scene reflectivity $\brho$ are non-negative real-valued numbers. Yet the optimization is performed in the complex domain with complex valued iterates. 
Given the limited number of layers in the network architecture, and desire to interpret the output of each layer as a visual representation of the scene, we form the representations by taking the absolute values of the complex valued iterates.  

Finally, using the proximity operators of $\ell_1$ and $\ell_0$ penalties \cite{combettes2009proximal, blumensath2008iterative}, we determine the network activation functions as 
\begin{eqnarray}\label{eq:l0_1_ProxOps}
\sigma^1_{\tau}(\brho) & = & \text{max} (|\brho| - \tau, 0 ) \\
\sigma^0_{\tau}(\brho) & = & \text{max} ( |\brho| - \tau^{0.5}, 0) + (1-c)\tau^{0.5} u(|\brho| - \tau^{0.5}) \nonumber
\end{eqnarray}
where $c>0$ is a small constant in the order of $1e-5$, and the superscripts $1$ and $0$ are associated with the $\ell_1$ and $\ell_0$ norm penalties, respectively. Note that the constant $c$ is set to $0$ in IHTA. However, in order to backpropagate threshold derivatives in the $\ell_0$ case, we set $c$ to a small non-zero constant (See Appendix \ref{Sec:Appendix} for details). Note that operations in (\ref{eq:l0_1_ProxOps}) are understood to be element wise.

By making every representation in the network an image estimate, the linear mapping and thresholding by (\ref{eq:l0_1_ProxOps}) can be interpreted as an enhancement operation in which the image estimate from the previous iterate is progressively enhanced in forward propagation with the learned parameters $\mathbf{Q}$ and $\mathbf{F}$ to better match the synthesized data to the true measurements.


Additionally, the scene reflectivity may be upper bounded given the operating frequencies of the receiver and typical scene refractive indices. Thus, without loss of generality, we assume that the scene reflectivity varies between 0 and 1 and normalize the final RNN output ${\brho}^L$, before projection onto the data space as follows:
\begin{equation}\label{eq:NormalizationFP}
{\brho}^* = \frac{{\brho}^L}{\ \ \| {\brho}^L \|_{\infty}}.
\end{equation}
This normalization of the final output enhances the effect of learning in light of the expected range of reflectivity values in the reconstructed image. 

\subsection{Backpropagation for Passive SAR}

\subsubsection{Incorporating Constraints into Learning}
In (\ref{eq:Mismatch}), we choose the mismatch function between $\F\brho^*_n$ and $\d_n$ as the $\ell_2$ norm squared function. The standard approach in addressing the resulting optimization problem is via the stochastic or batch gradient descent.
However, applying additive gradient descent updates would alter the mathematical structures of $\F$ and $\tau$ which we wish to preserve.
Therefore, we constrain the loss function, ${\mathcal{J}}_{D}$ with the properties of these parameters and modify it as follows:
\begin{equation}\label{eq:ModifiedLossConst}
\tilde{\mathcal{J}}_{D}(\F, \Q, \tau) = {\mathcal{J}}_{D}(\F, \Q, \tau ) + i_{C_{\mathbf{F}}}(\mathbf{F}) + i_{C_{\tau}}(\tau )
\end{equation}
where 
$i_{C_{\mathbf{F}}}$ and $i_{C_{\tau}}$ are the indicator functions of sets $C_{\mathbf{F}}$ and ${C_{\tau}}$ for $\mathbf{F}$ and $\tau$, respectively. 
The backpropagation is then performed by \emph{projected gradient descent} at each epoch and the updates resulting in (\ref{eq:updates}) is projected onto the feasible sets as follows:
\begin{eqnarray} \label{eq:UpdateParamConst}
\F^{l+1} & = & \mathcal{P}_{C_{\F}} (\F^{l} - \eta \nabla_\F\Jc_{D}[\theta^l]) \\
\tau^{l+1} & = & \mathcal{P}_{C_{\tau}} (\tau^{l} - \eta \partial_\tau\Jc_{D}[\theta^l]) \nonumber
\end{eqnarray}
where $\theta = \{\F,\Q,\tau\}$, $\theta^l$ denotes the $l^{th}$ iterate in backpropagation and $\mathcal{P}_{C_{\F}}$ and $\mathcal{P}_{C_{\tau}}$ are the projection operators for the set $C_{\F}$ and $C_{\tau}$, respectively.

The most significant structure we wish to impose is on $\mathbf{F}$. Under the small scene, short aperture and constant amplitude assumptions, the SAR forward model becomes a matrix of complex exponential entries with constant modulus. \footnote{This assumption is justified by having a broadband antenna at the receiver, a narrowband transmitted waveform with a flat spectrum, small scene and short aperture imaging geometry, all of which are likely to be satisfied for typical passive imaging scenarios.} The feasible set $C_{\mathbf{F}}$ then becomes the set of matrices with constant modulus entries, which has a closed form projection operator. 
This constraint leads to the following update equation for the entries of $\mathbf{F}$:
\begin{equation}\label{eq:FUpdate}
{\mathbf{F}}^{l+1}   =  \frac{\mathbf{F}^{l} -  \eta \nabla_{\mathbf{F}} \Jc_{D}[\theta^l]}{|\mathbf{F}^{l} -  \eta \nabla_{\mathbf{F}} \Jc_{D}[\theta^l]|}
\end{equation}
where $\nabla_{\mathbf{F}}$ is the gradient with respect to $\F$, $| \cdot |$ denotes element-wise absolute value operation and $\kappa$ stands for the known or estimated transmitter power. 

More elaborate constraints on $\F$ can be constructed if richer a priori information on the transmitter location or transmitted waveforms is available. For example, if approximate region including the transmitter is known, the feasible set $C$ can be built to include complex values whose phase and amplitude values are related to the locations of the cells in the region. Similarly, transmitted waveform related a priori information can be used to determine the phase and amplitude of complex values in the feasible set $C$.
The feasible set for the threshold value $\tau$ is positive real numbers, yielding the trivial projection:
\begin{equation}\label{eq:TauUpdate}
\tau^{l+1}  =  \text{max} (\tau^{l} - \eta \nabla_\tau\Jc_{D}[\theta^l] , 0 ).
\end{equation}
When approached from the optimization perspective, the image domain filter, $\Q$, is a high pass filter. Since $\F$ is already constrained, we choose not to constrain $\Q$. After being initialized as a ramp filter in spatial domain, learning $\Q$ and $\tau$ leads the network to a local minimum that ideally captures a superior denoising effect. However, a priori information on transmitted waveforms can be used to build specific constraints on $\Q$. 

\subsubsection{Network Derivatives}

The update equation in (\ref{eq:UpdateParamConst}) requires computation of the gradient of the mismatch function in (\ref{eq:Mismatch}) with respect to network parameters. 
We derive the complex gradients with respect to $\Q$ and $\F$ using \emph{Wirtinger calculus} \cite{candes2015phase} and determine the update equations using the theory of complex backpropagation \cite{leung1991complex}. 

Analytic expressions for the contribution of each layer to the gradient of the mismatch function with respect to $\F$ and $\Q$ as well as the derivative of $\ell$ with respect to $\tau$ are derived in Appendix \ref{Sec:Appendix} using tensor algebra and Wirtinger derivatives. We use the gradient contribution of the $k^{th}$ layer, $(\nabla_{\Q}\ell)^k$ and $(\nabla_{\F}\ell)^k$, and the definitions from (\ref{eq:Q_deriv_express}), (\ref{eq:F_ell_grad}) in the BPTT algorithm to compute the gradients of the loss function $\mathcal{J}_D$ as follows:
\begin{eqnarray}\label{eq:FinalGrad}
\nabla_{\mathbf{Q}} \mathcal{J}_D[\theta^l] & = & \frac{1}{T} \sum_{n=1}^T \sum_{k=1}^L (\nabla_{\mathbf{Q}} \ell(\d^*_n, \d_n))^k \big\vert_{\theta = \theta^l}  \\
\nabla_{\mathbf{F}} \mathcal{J}_D[\theta^l]  & = & \frac{1}{T} \sum_{n=1}^T  \big[ ({\d}^*_n - \d_n) {{\brho}^*_n}^T + \nonumber \\ 
& \ & \sum_{k=1}^L (\nabla_{\mathbf{F}} \ell(\d^*_n, \d_n))^k \big] \bigg\vert_{\theta = \theta^l} \nonumber \\
\partial_{\tau} \mathcal{J}_D[\theta^l]  & = & \frac{1}{T} \sum_{n=1}^T \sum_{k = 1}^L \sum_{i \in \mathit{I_k}} - (\nabla_{{\brho}^*_n} \ell( \d^*_n, \d_n))_i \big\vert_{\theta = \theta^l} \nonumber
\end{eqnarray}
where $\d^* = \F\brho^*(\F,\Q,\tau)$, the subscript $i$ denotes $i^{th}$ entry of the vector $\nabla_{{\brho}^*_n} \ell(\d^*_n, \d_n)$ 
 and $I_k$ denotes the index set for which $|\Q \brho^k + \alpha \F^H \d| > \tau$. All terms are evaluated at $\theta^l = \{\F^l, \Q^l, \tau^l \}$, where $l$ denotes the backpropagation update iteration.
Note that the gradient expressions for the $\ell_0$ and $\ell_1$ based constraints are identical up to a Dirac-delta term. Nevertheless, the two algorithms produce different results since the forward propagation steps are different. As a result the two algorithms result in different data mismatch and different backpropagation results.  

\subsection{Computational Complexity Analysis}

The computational complexity of backpropagation with analytically derived updates proves to be on comparable order to those of original ISTA and IHTA with $M$ number of unknowns and $N$ number of measurements. 

Component $\nabla_{{\brho}^*_n} \ell(\d^*_n, \d_n)$ arising from data mismatch and normalization steps is only computed once per sample at each epoch, resulting in $\mathcal{O}(M^2) + \mathcal{O}(NM)$ multiplications.
At each layer, for the $\Q$ and $\F$ derivatives in Eq. (\ref{eq:Q_deriv_express}),(\ref{eq:F_ell_grad}), we have in $\mathcal{O}(LM^2)$ and $\mathcal{O}(NM)$ computations from $\Q\brho^k$ and $\F^H\d$ multiplication operations respectively. 
The $M$ elements of vector $\Q \brho^k + \alpha \F^H \d$ are multiplied with the vectors $(\brho^{k-1})^T$ and $\d$ forming the $M$ rows and columns of $(\nabla_{\mathbf{Q}} \ell(\d^*_n, \d_n))^k$ and $(\nabla_{\mathbf{F}} \ell(\d^*_n, \d_n))^k$, respectively. These operations require $\mathcal{O}(LNM)$ and $\mathcal{O}(LM^2)$ computations. 
For $T$ training samples, and $E$ epochs, the overall computational complexity of backpropagation becomes $\mathcal{O}(ETLNM) + \mathcal{O}(ETL M^2)$.
The difference in computational complexity between DL-based inversion and that of standard iterative reconstruction depends on the ratio of $ETL$ to the number of iterations in standard method. 
\section{Numerical Simulations}\label{Sec:NumericalSim}
We perform numerical simulations to demonstrate the performance of DL based passive SAR imaging. We consider a typical passive SAR scenario and discuss how DL based approach can improve the results over conventional optimization methods. We consider the case in which the location or the look direction of the transmitter of opportunity is not known. We assess the performance of the DL-based approach with respect to two different figures of merit: image domain mismatch to ground truth and contrast of reconstructed image. Additionally, we consider the effects of regularization parameter initialization, network depth, and training data size in reconstructed images.

\subsection{Experimental Set-up}

\subsubsection{Scene and Imaging Parameters}\label{sec:sys_params}
\begin{figure}
\hspace*{-0.5cm}
\centering
\subfigure[]{
\includegraphics[width=0.28\textwidth]{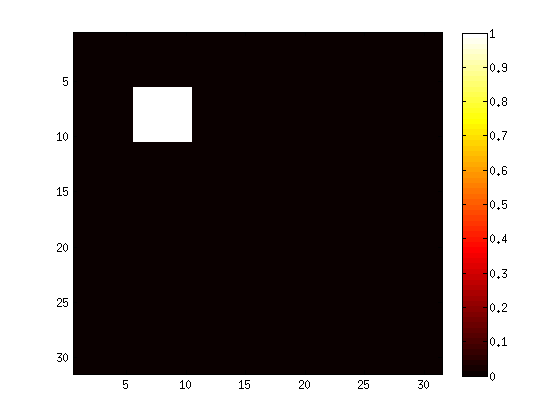}
\label{fig:test_img}
}%
\centering
\subfigure[]{
\includegraphics[width=0.28\textwidth]{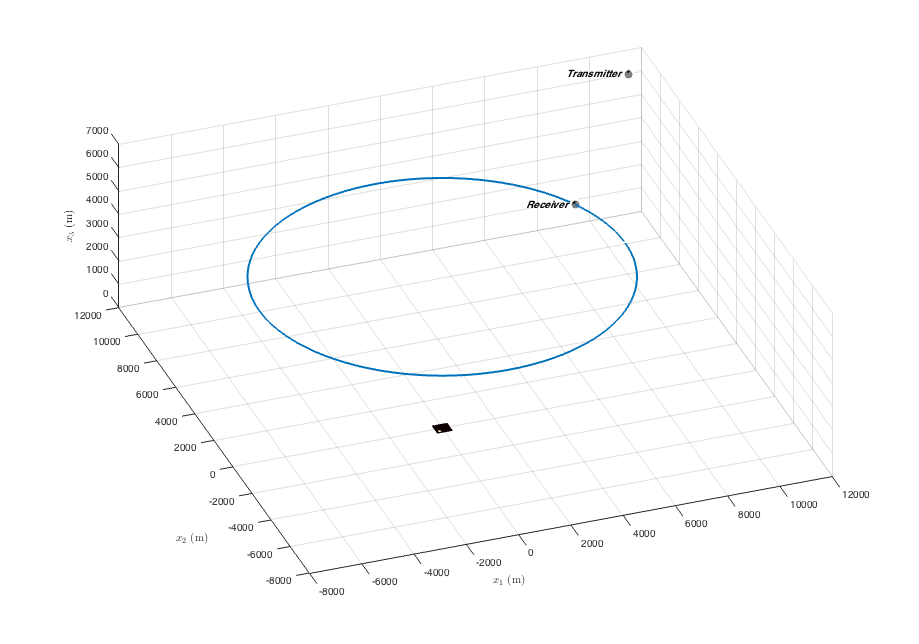}
\label{fig:scene_config}
}
\caption{Figs. \ref{fig:test_img}, \ref{fig:scene_config} display image phantom used to generate the test data set, and the imaging geometry used in simulations, respectively. The transmitter is placed approximately $15$\textmd{km} distanced from the center of the scene. The test set is generated by adding multiple realizations of Gaussian noise on the measurements obtained from the phantom scene.}
\end{figure}

We assume isotropic transmit and receive antennas, and simulate a transmitted waveform with a flat spectrum with bandwidth and center frequency of $8$MHz and $760$MHz, respectively.
The transmitted waveform parameters and properties correspond to a DVB-T signal, a commonly used illuminator of opportunity \cite{Palmer13}.
We create a flat scene that is $620\times 620$m$^2$ discretized into $31\times 31$ pixels with the origin of the coordinate system located at the center of the scene at pixel $(16,16)$. This results in each pixel having $20$m range resolution for monostatic SAR operating with $8$MHz bandwidth.  The receiver antenna traverses a circular trajectory, defined as $\bgamma_R(s) = [7 \cos(s), 7 \sin(s), 6.5]$km and the transmitter is fixed and located at $\bgamma_T = [11.2, 11.2, 6.5]$km. The received signal is generated using the bistatic forward model given in \eqref{eq:FIO_forward_model} with $A\equiv 1$.
The aperture is sampled by discretizing the slow-time variable into $400$ uniform samples, and the bandwidth is sampled by discretizing the fast-time variable into $100$ uniform samples.
This simulation configuration is displayed in Fig. \ref{fig:scene_config}.

\subsubsection{Training Data} \label{Ssec:TrainData}
We generate a training set consisting of sparse scenes with a single extended target that varies in rectangular shape and location. The length and width of each rectangular target is chosen randomly and lies in the range $[1,6]\times[1,6]$ pixels. The targets are placed randomly within the range of $[3,28]\times[3,28]$ pixels.
With this approach, the locations and size of the targets in our training set are all realizations from the same uniform distribution. The possible training images correspond to scenes with point and extended targets.
We then generate SAR data for each image using the full forward model described in Section \ref{sec:sys_params}.

 We envision a two-stage data collection protocol to collect training and testing data for use with DL based passive SAR imaging. In the first stage an airborne receiver collects test data from a scene of interest. In the second stage, several reflectors are placed in the scene to form either extended or point targets and training data is collected under the same imaging geometry as before. 

Another protocol is to collect the training data set over the course of an extended period during which temporary structures may appear creating perturbations in the background scene of interest. 
Note that neither the location nor the shape of the foreground scatterers need to be known in order to use the training data in our unsupervised training scheme.



\subsubsection{Testing Data}\label{Ssec:TestData}

While in the training process we create and use a set of varying images to train the network and tune the parameters, in testing we use data collected from a single scene of interest. The scene of interest is displayed in Fig. \ref{fig:test_img} and backscattered field is generated by the bistatic forward operator in Section \ref{sec:sys_params}. The data includes backscattered field embedded in additive white Gaussian noise at an SNR of $50$dB. 20 different realizations of noise is used to form a set of 20 images. The figures of merit are averaged over 20 results to obtain a statistical evaluation.

\subsubsection{Network Architecture, Initialization and Learning}
We use 16 layers for the RNN encoder with the activation functions discussed in Section \ref{Sec:DLPassive}. Training is performed with projected batch gradient descent in which the gradient is averaged over all training samples before being projected onto the sets defined by constraints described in Section \ref{Sec:SARDL}. In our simulations, the model was trained for $7$ epochs at which point, in most cases, we observed the backpropagation converged. To help ensure convergence of the training procedure, we down scale the step size of gradient descent at a rate of $\eta_l = \eta_l /(1 + l)$ where $l = 0,1,\dots$ indexes the epoch.
In our simulations we tune the learning rates differently for each parameter as $1e-9$ for $\Q$, $1e-5$ for $\F$, and $1e-14$ for $\tau$. 

Unlike typical deep learning applications where parameters are initialized randomly, in our method, the parameters are initialized with the partially known forward model.
Since the transmitter location is unknown in passive SAR imaging, the parameter $\F$ is initialized with the finite dimensional version of the operator $\Fc$, in which the range term is $|\bgamma_R(s)-\x| + \phi_T(s,\bi x)$.
When the location of the transmitter is not known, we set $\phi_T(s,\bi x) = 0$. When partial knowledge about the transmitter location is available, $\phi_T(s,\bi x)$ can be chosen accordingly.
We use the initial value of $\mathbf{F}$ to initialize the filter parameter $\Q$ and bias $\mathbf{b}$ according to \eqref{eq:WeightM}, where $\alpha=1e-6$, upper bounded by the largest eigenvalue of $\F^H \F$. This choice of $\alpha$ holds for system parameters discussed above, and in the case of ISTA and IHTA ensures monotonic descent of the objective function. The $\alpha$ parameter is kept fixed in all experiments as it is learned within the process of learning $\mathbf{Q}$ and $\mathbf{F}$.

\subsubsection{Figures of Merit}\label{Sssec:Figmerit}
In training, we use the normalized $\ell_2$ error in data domain at each epoch as a stopping criterion:
\begin{equation}\label{eq:dataMismatch}
L_{\d}( \mathcal{\brho^*}^l )= \frac{\| \F^l (\mathcal{\brho^*})^l - \d \|_2^2}{\| \d \|_2^2}
\end{equation}
where $(\brho^*)^l$ is the image reconstructed by the encoder in epoch $l$, $\F^{l}$ is the learned forward model at epoch $l$, and $\d$ is the input data. Training is terminated once (\ref{eq:dataMismatch}) does not decrease in average. Final learned parameters are set from the epoch that produced the lowest average $L_{\d}$ over the training set.

To quantify the performance of DL based image reconstruction, we consider two figures of merit: normalized $\ell_2$ error in image domain $L_{\brho}$ with respect to the ground truth, and contrast measure $C_{\brho}$ defined as: 
\begin{equation}\label{eq:Merits}
L_{\brho}( \mathcal{\brho^*}) = \frac{\| \mathcal{\brho^*} - \brho \|_2^2}{\| \brho \|_2^2}, \quad C_{\brho}({\brho^*}) = \frac{|\text{E}[{\brho^*}_f] - \text{E}[\brho^*_b]|^2}{\text{var}[{\brho^*}_b] }
\end{equation}
where $\brho^*$ denotes the reconstructed image, $\brho^*_{f}$ and $\brho^*_{b}$ are the foreground and background images, respectively, $\text{E}$ is statistical expectation, and $\text{var}$ is statistical variance.

In our experiments, we investigate the impact of initialization of $\lambda$ parameter, the size of the training set and the number of layers of the network.
The images reconstructed by $16$-layer network are compared to the results of ISTA and IHTA obtained after $100$ iterations.

\subsection{Results}



In this section we present and compare the performance of DL based passive SAR to those of standard IHTA and ISTA.
We demonstrate that the learned networks produce higher quality imagery with far fewer iterations.

\subsubsection{Impact of $\lambda$ Initialization}\label{sec:lambda_vary}

\begin{figure}[!ht]
\centering
\subfigure[]{
\includegraphics[width=0.35\textwidth]{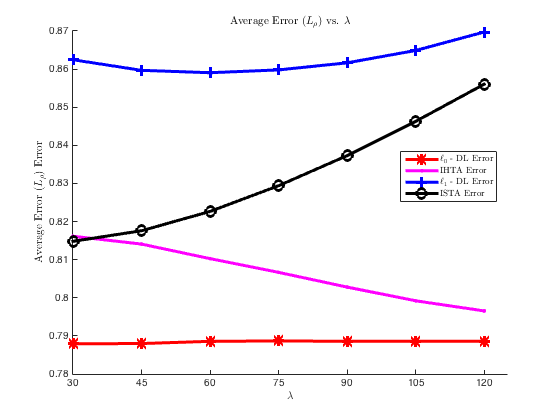}
\label{fig:lambdavsError}
}
\centering
\subfigure[]{
\includegraphics[width=0.35\textwidth]{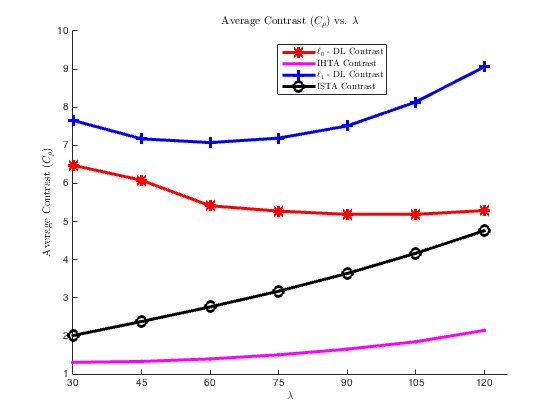}
\label{fig:IHTA_lambdavsContrast}
}
\caption{Image domain $\ell_2$ \subref{fig:lambdavsError} error and contrast \subref{fig:IHTA_lambdavsContrast} for different values of $\lambda$ initialization produced by the $16$-layer $\ell_0$-DL (red) and $\ell_1$-DL (blue) method, generated after $7$ epochs of training, and $100$ iterations of standard IHTA (magenta) and ISTA (black line). Contrast values are plotted in $\text{log}$ scale. }
\label{fig:lambda_QuatitativeComp}
\end{figure}

We train the proposed 16-layer network with initial regularization parameters varying as $\lambda = 30, 45, 60, 75, 90, 105, 120$, and plot the results obtained for different initial $\lambda$ value for the figure of merits given in \eqref{eq:Merits}, in the left and right subfigures of Fig. \ref{fig:lambda_QuatitativeComp}, respectively.

\begin{figure*}[!ht]
\hspace*{-1.5cm}
\centering
\subfigure[]{
\includegraphics[width=0.28\textwidth]{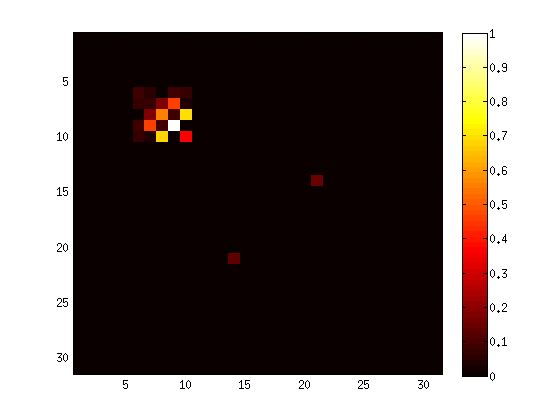}
\label{fig:DL_ihta_l30_max_contrast}
}%
\centering
\subfigure[]{
\includegraphics[width=0.28\textwidth]{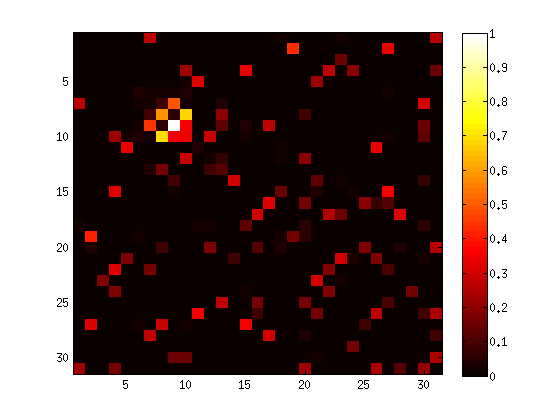}
\label{fig:ihta_l120_max_contrast}
}%
\centering
\subfigure[]{
\includegraphics[width=0.28\textwidth]{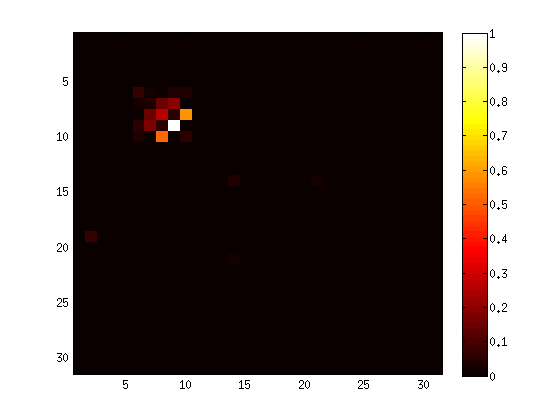}
\label{fig:DL_ista_l120_max_contrast}
}%
\centering
\subfigure[]{
\includegraphics[width=0.28\textwidth]{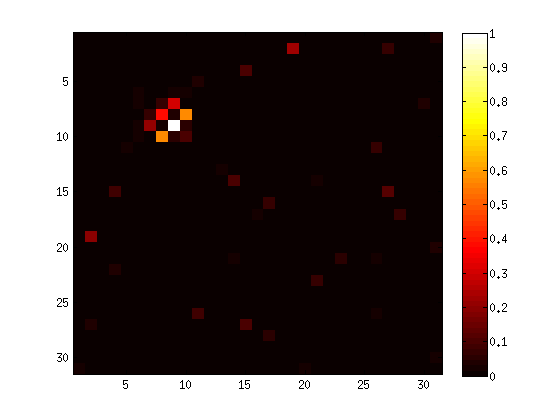}
\label{fig:ista_l120_max_contrast}
}
\caption{Images generated using the $\lambda$ values that provide the maximum reconstruction contrast for ISTA and IHTA inspired DL and the corresponding standard algorithms for $100$ iterations. \subref{fig:DL_ihta_l30_max_contrast} The image formed by the IHTA inspired DL network with $\lambda=30$. \subref{fig:ihta_l120_max_contrast} The image formed by standard IHTA with $\lambda=120$. \subref{fig:DL_ista_l120_max_contrast} The image formed by ISTA inspired DL with $\lambda=120$. \subref{fig:ista_l120_max_contrast} The image formed using standard ISTA with $\lambda = 120$.}
\label{fig:best_images}
\end{figure*}

Fig. \ref{fig:best_images} shows reconstructed images using $\lambda$ values that result in the maximum image contrast for the 16-layer $\ell_0$ and $\ell_1$-DL, and for $100$ iterations of the corresponding standard algorithms.
Visually it is clear that DL method yields significant improvement in contrast, and comparable image domain $\ell_2$ error at only $16$ layers.

In $\ell_0$ regularized case, for both metrics we see that the DL method outperforms the IHTA algorithm, specifically in terms of background suppression and superior geometrically fidelity of the target when observed in Fig. \ref{fig:best_images}. 
For $\ell_1$ regularized case, compared to ISTA, DL based approach significantly enhances the contrast measure of images, however, the inverse correlation of $\ell_2$ image domain error and contrast is lost, as the $\ell_1$-DL method yields a higher average $\ell_2$ error over the test set. 
This behavior can be attributed to significant suppression of target pixels. 
Even though the background scatterers are effectively suppressed to boost contrast measure, the mean of foreground pixels is also considerably decreased with the $\ell_1$-DL method, causing the second moment in the foreground to dominate the variance, resulting with an increase in both contrast and $\ell_2$ image error metrics simultaneously.
The improved geometric fidelity can be primarily explained by the refinement of the forward model with learning, as $\F$ is the component that determines where the target is placed via its adjoint. 
Superior background suppression in DL based images is a direct result of the learned threshold value and image domain filter, which together serve as denoising operators on representations in the network.


An empirical detail in setting the initial $\lambda$ is preventing strong suppression of pixel values in the first forward propagation. 
A large initial threshold value would produce highly sparse representations in the network, resulting with backpropagating considerably smaller magnitudes for parameter updates. 
The improvement of DL over conventional methods in such scenario diminishes significantly.
In our experiments, we observed $\lambda \geq 150$ to be the point where the performance deteriorates for our configuration, hence we limit the range of $\lambda$ upto $120$ in our experiments.

\subsubsection{Impact of Training Set Size}

\begin{figure}
\centering
\subfigure[]{
\includegraphics[width=0.35\textwidth]{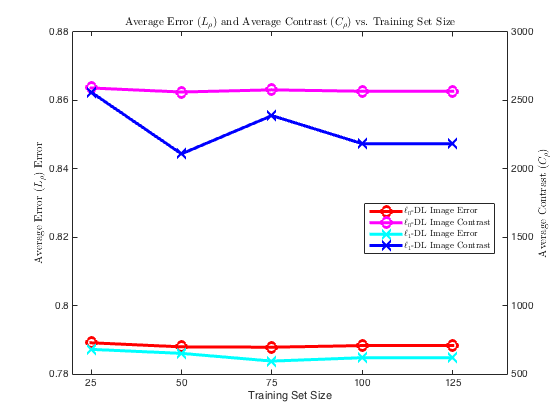}
\label{fig:training_size_metric}
}
\centering
\subfigure[]{
\includegraphics[width=0.35\textwidth]{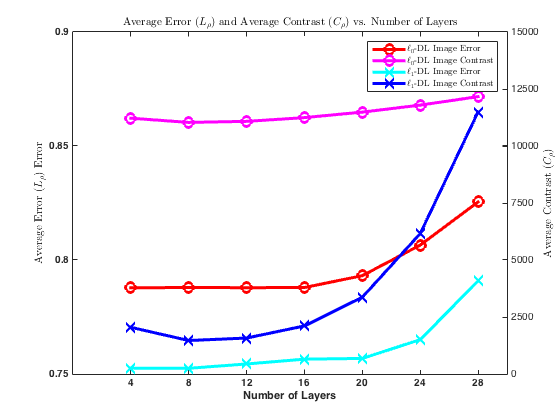}
\label{fig:layer_number_metric}
}
\caption{The reconstruction error (left axis) and contrast (right axis) vs. number of training samples in \subref{fig:training_size_metric}, vs. number of layers in \subref{fig:layer_number_metric} for $\ell_0$-DL and $\ell_1$-DL methods. In both figures, image domain error and contrast curves for $\ell_0$-DL are in red and magenta, for $\ell_1$-DL are in cyan and blue respectively.}
\label{fig:QuatitativeComp}
\end{figure}

In this set of experiments we fix $\lambda = 30$ and train our model with varying training set sizes of $T = 25, 50, 75, 100, 125$.
In Fig. \ref{fig:training_size_metric}, we plot the performance measured by the figure of merits given in \eqref{eq:Merits} vs. the training set size.
Our results indicate that increasing the training set size from $25$ up to $125$ does not provide any significant improvement in either metric.
The reason for this negligible effect on performance can be primarily explained by our initialization. As the network is already initialized with a known component of the forward model, the physical process mapping the scene to data is established to some degree. Our experiment suggests that despite the size of the training set, backpropagation converges to a very similar local minimizer.

A concern with small training sets is the risk of overfitting parameters.
Especially with our protocol discussed in Section \ref{Ssec:TrainData}, the training data is likely to be correlated with test data.
While such correlation between training and testing is most likely unavoidable in imaging problems, our experiment suggests that this is not a problem. Furthermore, correlation may lead to benefits such as mitigating the impact of using a larger training set, hence reduce the computational complexity of training significantly. 

\subsubsection{Effect of Network Depth}

Finally, we investigate the impact of number of layers $L$ in designing the deep network for DL based reconstruction. We measure performance quality with our two figures of merit and plot the results obtained from models trained with $L = 4, 8, 12, 16, 20, 24, 28, 32$ layers, trained with $50$ samples, keeping $\lambda = 30$ for both cases. Figure \ref{fig:layer_number_metric} displays the image domain mismatch and contrast metrics vs. $L$ in left axis and right axis of the plot respectively.

With Figure \ref{fig:layer_number_metric}, we clearly observe that there is a trade-off between reconstruction error and image contrast as number of layers increase.
This behavior can best be explained by the increase number of filtering and thresholding operations applied to the backprojected input data with more layers.
With more successive denoising operations in the network, components, such as the corners of the target become more prone to suppression.
Background suppression is also enhanced with increased layers which yields images with superior contrast. 
Furthermore, an increase in layers suggests that derivatives are summed over more stages with BPTT algorithm, this accelerates learning of $\Q$ and $\tau$ more than $\mathbf{F}$, since the forward model is subject to a unit modulus constraint.
We suspect this causes more target suppression effects in the final image estimate.

As the number of layers decrease, the reconstruction error decreases due to superior matching of the target.
Since there is better match of large pixel values there is a decrease in the mismatch error, however background suppression is less effective, limiting the contrast performance.
Therefore, the number of layers serves as a tunable hyper-parameter to determine the trade-off between target visualization, background suppression, and computational complexity.

\section{Conclusion}\label{Sec:Concl}

We present a novel deep learning based approach to inverse problems in imaging and demonstrate its application in passive SAR imaging. 

We review basic concepts and tools in DL and consider forward modeling and inversion for imaging in DL framework. In the same way DL extends conventional machine learning, we extend conventional forward modeling and image reconstruction by inserting additional layers in between conventional two-layer forward and inverse solvers. DL based forward modeling is capable of capturing non-linearities between physical measurements and quantity of interest more accurately than its conventional linear counterparts. Similarly, we extend the conventional two-layer image reconstruction method by inserting additional hidden layers in between backprojection and filtering. We motivate the choice of network activation function from a Bayesian formulation of image reconstruction with non-Gaussian prior models and associated numerical optimization algorithms.  Speciﬁcally, we design a RNN architecture as an inverse solver based on the iterations of proximal gradient descent optimization methods. We further adapt the RNN architecture to image reconstruction problems by adding an additional layer transforming the network into a recurrent auto-encoder. Our DL based inverse solver is particularly suitable for a class of image formation problems in which the forward model is only partially known. The ability to learn forward models, hyper parameters combined with unsupervised training approach establish our recurrent auto-encoder suitable for real world applications.

Unlike existing methods for passive imaging, our approach only requires a single receiver and can image arbitrary scenes with no knowledge of the transmitter location or transmitted waveform, making our method ideal for contested environments.
We initialize the network with the partially known forward operator, providing a good initialization for the highly non-convex training optimization problem.
With unsupervised training, the forward model and image filters are refined to compensate for the missing phase information, which drives the model to form more accurate imagery in forward propagation, foregoing the performance of limitations of conventional imaging algorithms.
We provide extensive numerical simulations to demonstrate the power of the DL based approach using networks based on network activation functions derived from $\ell_1$ and $\ell_0$ prior models.
Our experiments show that the learned network reconstructs images with better geometric fidelity, higher contrast and reduced reconstruction error, requiring significantly fewer iterations than those of conventional ISTA and IHTA. Additionally, we show that the DL based method is comparable in terms of complexity to running conventional least-squares based optimization problems for a large number of iterations.

\bibliographystyle{IEEEtran}
\bibliography{IEEEabrv,ref}
\appendix \label{Sec:Appendix}
\subsection{Backpropagation Derivatives}
We compute the gradient contributions $(\nabla_{\F} \ell)^k$, $(\nabla_{\Q} \ell)^k$ from $k^{th}$ layer of the network to sum over $k$ with the \emph{backpropagation through time}(BPTT) algorithm, and $\partial_{\tau} \ell$, for update equations (\ref{eq:updates}). 
Due to having real valued representations such that $\brho^k \in \mathbb{R}^M$, and $\bar{\brho}^* = {\brho}^*$ the complex backpropagation equation becomes:
\begin{eqnarray}\label{eq:FinalNetBackprop}
\frac{\partial \ell(\d^*, \d)}{\partial \mathbf{Q}} & = & \frac{\partial \brho^*}{\partial \mathbf{Q}} \big( \mathbf{F}^T (\bar{\d^*} - \bar{\d}) + \mathbf{F}^H ({\d}^* - \d) \big) \\
\frac{\partial \ell(\d^*, \d)}{\partial \mathbf{F}} & = & \frac{\partial {\brho}^*}{\partial \mathbf{F}}\big(  \mathbf{F}^T (\bar{\d^*} - \bar{\d}) + \mathbf{F}^H (\d^* - \d) \big) \nonumber \\ 
& \ & \quad + \frac{\partial \mathbf{F}}{\partial \mathbf{F}} \brho^* (\bar{\d^*} - \bar{\d}) \nonumber \\
\frac{\partial \ell(\d^*, \d)}{\partial \tau} & = & \frac{\partial {\brho}^*}{\partial \tau} \big( \mathbf{F}^T (\bar{\d^*} - \bar{\d}) + \mathbf{F}^H ({\d}^* - \d) \big) \nonumber
\end{eqnarray}
From BPTT algorithm, derivatives of the normalized image output $\brho^*$ from equation (\ref{eq:FinalNetBackprop}) can be written as:
\begin{equation}\label{eq:BPTTRnn}
\frac{\partial \ell(\d^*, \d)}{\partial \theta} = \left(\sum_{i = 1}^{L} \frac{\partial \brho^k}{\partial \theta}\right) \nabla_{{\brho}^*} \ell(\d^*, \d)
\end{equation}
where $\theta$ is a surrogate for network parameters $\{\mathbf{Q}, \mathbf{F}, \tau \}$, $\brho^k$ is the representation at the $k$th layer, and
\begin{equation}\label{eq:DefNablRhostar}
\nabla_{{\brho}^*} \ell(\d^*, \d) = \left(- \frac{1}{\|{\brho}^L \|_{\infty}^2} \frac{\partial \|{\brho}^L\|_{\infty}}{\partial {\brho}^L} {\brho^L}^T + \frac{1}{\|{\brho}^L\|_{\infty}} \mathbf{I}_{M\times M}\right)$$
$$
\quad \times 2 \text{ Re}\{\mathbf{F}^H ({\mathbf{d}}^* - \mathbf{d})\}
\end{equation}
The first term of the multiplication is derived from the normalization derivative $\frac{\partial {\brho}^*}{\partial {\brho}^L}$. 
Denoting the argument of the threshold functions as $\mathbf{f}^{k+1} =  |\mathbf{Q} \brho^{k} + \alpha \mathbf{F}|$, we have that $\frac{\partial {\brho}^k}{\partial \theta} = \frac{\partial \mathbf{f}^k}{\partial \theta} \frac{\partial \brho^k}{\partial \mathbf{f}^k}$. $\frac{\partial \brho^k}{\partial \mathbf{f}^k}$ is merely the derivative of the thresholding function $\mathcal{P}_{\alpha \lambda \Phi}(\cdot)$, which equals a diagonal matrix with entries $1$ at indexes that $\mathbf{f}^k_i> \tau$, and equals $0$ otherwise. 

\subsubsection{$\mathbf{Q}$-Gradient}

Letting $f^k_i = (z^{k}_i \bar{(z^{k}_i)})^{\frac{1}{2}}$, we have $\frac{\partial \mathbf{f}^k_i}{\partial \mathbf{Q}} = \frac{\partial z^k_i}{\partial \mathbf{Q}} \frac{\partial \mathbf{f}^k_i}{ \partial z^{k}_i}$, where $z^k_i = \Q_i \brho^k + \alpha (\F^H)_i \d$, $i$ subscript denoting the $i^{th}$ row of matrices $\Q$ and $\F^H$. 
Since $\frac{\partial \rho^k}{\partial \mathbf{f}^{k}}$ is merely a diagonal matrix with diagonal elements $\{0, 1\}$, such that it equals $1$ if $ \mathbf{f}^{k}_i > \tau$, its essentially a selection matrix for the entries of $\nabla_{{\brho}^*} \ell$, suppressing indexes that fall under the threshold. Following with a tensor-vector multiplication, the derivative becomes:
\begin{equation}\label{eq:Q_deriv_TM}
\frac{\partial \ell(\d^*, \d)}{\partial \mathbf{Q}} = \sum_{k=1}^L  \sum_{i = I_{\mathbf{f}^k}} \frac{\partial \mathbf{f}^{k}_i}{\partial \mathbf{Q}} (\nabla_{{\brho}^*} \ell)_i
\end{equation}
where $I_{\mathbf{f}^k}$ is the set of indices that $\mathbf{f}_i^k > \tau$ is satisfied. For each $k = 1, \cdots L$, from the definition of complex gradient operator, for each row $i = 1, \cdots M $ we obtain:
\begin{equation}\label{eq:Q_deriv_express}
(\nabla_{\mathbf{Q}} \ell)_{i,:}^k = \frac{(\nabla_{{\brho}^*} \ell)_i}{2}  \frac{\mathbf{Q}_i \brho^k + \alpha \mathbf{F}^H_i \d}{|\mathbf{Q}_i \brho^k + \alpha \mathbf{F}^H_i \d|} (\brho^{k-1})^T 
\end{equation}
if $|\mathbf{Q}_i \brho^k + \alpha \mathbf{F}^H_i \d| > \tau$ and $0$'s everywhere else. 

\subsubsection{$\mathbf{F}$-Derivative}
For the second expression from equation (\ref{eq:FinalNetBackprop}),  we obtain $\left(\frac{\partial \mathbf{F}}{\partial \mathbf{F}} {\brho}^*\right) (\bar{\d}^* - \bar{\d}) = (\bar{\d}^* - \bar{\d}) {\brho^*}^T$.
The first term in equation (\ref{eq:FinalNetBackprop}) for $\F$ is identical to $\mathbf{Q}$-derivative in equation (\ref{eq:Q_deriv_TM}) except for the variable at differentiation of $\mathbf{f}^k$. Following the same steps as in differentiation with respect to $\mathbf{Q}$, for each row $i = 1, \cdots M$, we obtain:
\begin{equation}\label{eq:F_ell_grad}
(\nabla_{\mathbf{F}} \ell)_{:,i}^k = \frac{\alpha (\nabla_{{\brho}^*} \ell)_i}{2}  \frac{\overline{(\mathbf{Q}_i \brho^k + \alpha \mathbf{F}^H_i \d)}}{|\mathbf{Q}_i \brho^k + \alpha \mathbf{F}^H_i \d|} \d
\end{equation}
if $|\mathbf{Q}_i \brho^k + \alpha \mathbf{F}^H_i \d| > \tau$ and $0$ everywhere else. 

\subsubsection{$\tau$-Derivative}

Since all inputs to activation function are positive $\brho^k_i = \max (0, \mathbf{f}^k_i - \tau)$, the derivative $(\frac{\partial \brho^k}{\partial \tau})_{1 \times N}$ will equal $-1$ at indexes $\mathbf{f}^k_i > \tau$ and $0$ otherwise. Then the from $k$th layer derivative becomes:
\begin{equation}\label{eq:Tau_deriv}
\partial^1_{\tau} \ell = \sum_{k = 1}^L \sum_{i \in \mathit{I_k}} - (\nabla_{{\brho}^*} \ell)_i
\end{equation}
where $\mathit{I_k}$ is the set of indexes where $|\mathbf{Q}_i \brho^k + \alpha \mathbf{F}^H_i \d|  > \tau$, superscript $1$ indicating the derivative for soft threshold function. For the hard thresholding operator with $c=0$ in (\ref{eq:l0_1_ProxOps}), the $\tau$ derivative would have the index set $\mathit{I_k}$ as indexes $i$ where $|\mathbf{Q}_i \brho^k + \alpha \mathbf{F}^H_i \d|  = \tau$. Due to double precision in computations, exact equality is of negligible probability, which would suppress the threshold derivative in backpropagation. Once $c = 1e-5$ is inserted into the function definition, the $\tau$ derivative in hard thresholding becomes $\partial^0_{\tau} \ell = (1e-5) \partial^1_{\tau} \ell$.

\end{document}